# Leaf Classification Using Shape, Color, and Texture Features

Abdul Kadir[#1], Lukito Edi Nugroho[*2], Adhi Susanto[#3], Paulus Insap Santosa[#4]

*Department of Electrical Engineering, Gadjah Mada University*
*Yogyakarta, Indonesia*

*Abstract*— **Several methods to identify plants have been proposed by several researchers. Commonly, the methods did not capture color information, because color was not recognized as an important aspect to the identification. In this research, shape and vein, color, and texture features were incorporated to classify a leaf. In this case, a neural network called Probabilistic Neural network (PNN) was used as a classifier. The experimental result shows that the method for classification gives average accuracy of 93.75% when it was tested on Flavia dataset, that contains 32 kinds of plant leaves. It means that the method gives better performance compared to the original work.**

## I. INTRODUCTION

Plant identification systems have been performed by several researchers. Wu et al. [1] identified 6 species of plants. They used aspect ratio, leaf dent, leaf vein, and invariant moment to identify plant. Wu et al. [2] proposed an efficient algorithm for plant classification. They involved 32 kinds of plants. Several features such as aspect ratio (ratio between length and width of leaf), ratio of perimeter to diameter of leaf, and vein features were used to characterize the leaf with accuracy of 90,312%. They also shared their data, called Flavia dataset, for academic research purpose. That dataset was used by Singh et al. [3] that did a research to compare Wu's algorithm to other methods: Support Vector Machine (SVM) and Fourier moment. Du et al. [4] captured the leaf shape polygonial approximation and algorithm called MDP (modified dynamic programming) for shape matching. However, all mentioned researchers did not incorporate color information in their identification systems.

Actually, shape, color and texture features are common features involved in several applications, such as in [5] and [6]. However, some researchers used part of those features only. Invariant moments proposed by Hue [7] are very popular in image processing to recognize objects [8] [9], including leaves of plants. Zulkifli [10] used invariant moments and General Regression Neural Network. Zulkifli worked on 10 kinds of leaves and did not process color information. However, according to our work [11], Polar Fourier Transform (PFT) proposed by Zhang [12] is better than invariant moments. Color was included in several applications as features, for example in [13], which used image correlogram for image retrieval, and in [14] that used color moments for plant classification.

According to Choras [15], texture is a powerful regional descriptor that helps in retrieval process. Texture, on its own does not have the capability of finding similar images, but it can be used to classify textured images from non-textured ones and then be combined with another visual attribute like color to make the retrieval more effective. Texture features can be extracted by using various methods. Gray-level occurrence matrices (GLCMs), Gabor Filter, and Local binary pattern (LBP) are examples of popular methods to extract texture features. Other method to get texture features is using fractals. Fractals in texture classification have been discussed in [16]. Fractal application in image retrieval has been applied by Min et al. [17]. There is a fractal measure called lacunarity, which is a measure of non-homogeneity of the data [18]. It measures lumpiness of the data. It defined in term of the ratio of the variance over the mean value of the function. One of lacunarity's definitions is used in this research, by some modifications in application.

Several identification systems used a neural network as a classifier. Neural networks have been attracted researchers in area pattern recognition because its power to learn from training dataset [8]. For example, back-propagation was used in [19] for adaptive route selection policy in mobile ad hoc networks. PNN is another neural network that has been used in several applications, such as in [20] for remote sensing image segmentation and in [21] for surface defect identification. According to [21], PNN has proven to be more time efficient than conventional back-propagation based networks and has been recognized as an alternative in real-time classification problems.

In this research, we tried to capture shape, color, vein, and texture of the leaf. In implementation, we used Fourier descriptors of PFT, three kinds of geometrics features, color moments, vein features, and texture features based on lacunarity. Then, those features were inputted into the identification system that uses a PNN classifier. Testing was done by using Flavia dataset. The result shows that the method improves performance of the identification system compared to Wu's result [2].

The remainder is organized as follows: Section 2 describes all features used in the research, Section 3 explains how the mechanism of experiments is accomplished, Section 4 presents the experimental results, and Section 5 concludes the results.

## II. FEATURE EXTRACTION

In the research, the features were extracted from shape, color, vein, and texture of the leaf. All features were used in identification system is described as follows.

### A. Shape Features

Two kinds of shape features used in the identification system are geometric features and Fourier descriptors of PFT. Geometric features that commonly used in leaves recognition are slimness and roundness. Slimness (sometime called as aspect ratio) is defined as follow:

$$slimness = \frac{l_1}{l_2} \qquad (1)$$

where $l_1$ is the width of a leaf and $l_2$ is the length of a leaf (Fig.1).

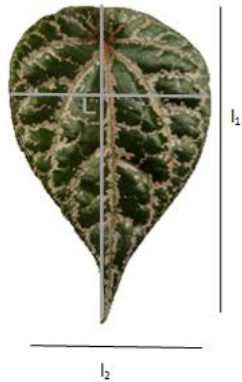

Fig. 1 Parameters for slimness of leaf

Roundness (or compactness) is a feature defined as:

$$roundness = \frac{4\pi A}{P^2} \qquad (2)$$

where A is the area of the leaf image and P is the perimeter of the leaf contour.

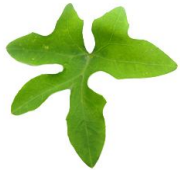

Fig. 2 Leaf with irregular shape

Dispersion (irregularity) is another feature suggested by Nixon & Aguado [22] to deal with an object that has irregular shape such as the leaf in Fig. 2. This feature is defined as

$$dispersion = \frac{\max(\sqrt{(x_i - \bar{x})^2 + (y_i - \bar{y})^2})}{\min(\sqrt{(x_i - \bar{x})^2 + (y_i - \bar{y})^2})} \qquad (3)$$

where $(\bar{x}, \bar{y})$ is the centroid of the leaf, and $(x_i, y_i)$ is the coordinate of a pixel in the leaf contour.

The Eq. 3 defines the ratio between the radius of the maximum circle enclosing the region and the minimum circle that can be contained in the region. Therefore, the measure will increase as the region spreads. However, dispersion has a disadvantage. It is insensitive to slight discontinuity in the shape, such as a crack in a leaf [22].

Polar Fourier Transform (PFT) are very useful to capture shape of a leaf. The descriptors extracted from PFT are invariant under the actions of translation, scaling, and rotation as illustrated in Fig. 3.

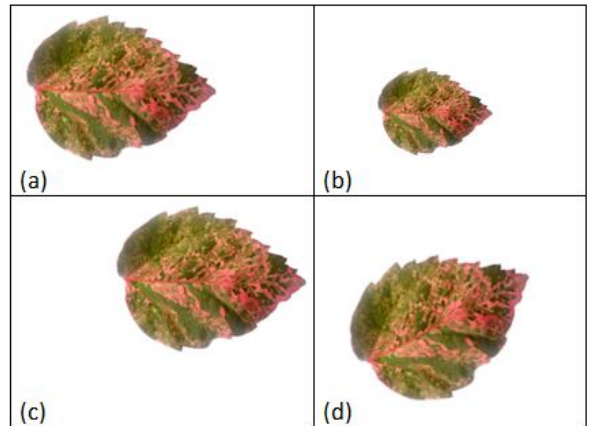

Fig. 3 Translation, scaling, and rotation invariants (a) leaf, (b) change of size, (c) change of position, (d) change of orientation

PFT that was used in this research is defined as

$$PF2(\rho, \phi) = \sum_r \sum_i f(\rho, \phi_i) \exp[j2\pi(\frac{r}{R}\rho + \frac{2\pi}{T}\phi)] \qquad (4)$$

where

- $0 \le r < R$ dan $\theta_i = i(2\pi/T)$ $(0 \le i < T); 0 \le \rho < R$, $0 \le \phi < T$,
- R is radial frequency resolution,
- T is angular frequency resolution.

Computation of PFT is described as follow. For example, there is an image I = {f(x, y); $0 \le x < M$, $0 \le y < N$}. Firstly, the image is converted from Cartesian space to polar space $I_p$ = {f(r,θ); $0 \le r < R$, $0 \le \theta < 2\pi$ }, where R is the maximum radius from the centre of the shape. The origin of polar space becomes as centre of space to get translation invariant. The centroid $(x_c, y_c)$ calculated by using formula:

$$x_c = \frac{1}{M}\sum_{x=0}^{M-1}x, y_c = \frac{1}{N}\sum_{x=0}^{N-1}y, \qquad (5)$$

In this case, (r, □) is computed by using:

$$r = \sqrt{(x-x_c)^2 + (y-y_c)^2}, \theta = \arctan\frac{y-y_c}{x-x_c} \qquad (6)$$

Rotation invariance is achieved by ignoring the phase information in the coefficient. Consequently, only the magnitudes of coefficients are retained. Meanwhile, to get the scale invariance, the first magnitude value is normalized by the area of the circle and all the magnitude values are normalized by the magnitude of the first coefficient. So, the Fourier descriptors are:

$$FDs = \{\frac{PF(0,0)}{2\pi^2}, \frac{PF(0,1)}{PF(0,0)}, ..., \frac{PF(0,n)}{PF(0,0)}, ..., \frac{PF(m,0)}{PF(0,0)}, ..., \frac{PF(m,n)}{PF(0,0)}\} \qquad (7)$$

where m is the maximum number of the radial frequencies and n is the maximum number of angular frequencies.

Fig. 4 shows two leaves and Table 1 lists Fourier descriptors of both leaves, using m=4 and n=6.

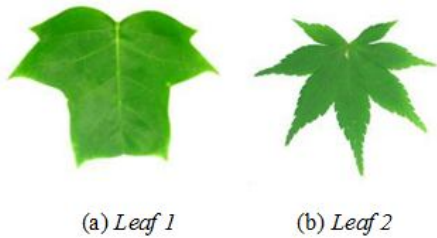

(a) *Leaf 1*        (b) *Leaf 2*

Fig. 4 Two kinds of leaves

TABLE 1
PART OF FOURIER DESCRIPTORS

| Features | Leaf 1 | Leaf 2 |
|---|---|---|
| 1 | 0.5591 | 0.3727 |
| 2 | 0.0024 | 0.0169 |
| 3 | 0.0412 | 0.0848 |
| 4 | 0.1809 | 0.0655 |
| 5 | 0.0187 | 0.0264 |
| 6 | 0.0768 | 0.0954 |
| 7 | 0.4155 | 0.4634 |

### B. Color Features

Color moments represent color features to characterize a color image. Features can be involved are mean ($\mu$), standard deviation ($\sigma$), skewness ($\theta$), and kurtosis ($\nu$). For RGB color space, the three features are extracted from each plane R, G, and B. The formulas to capture those moments:

$$\mu = \frac{1}{MN}\sum_{i=1}^{M}\sum_{j=1}^{N}P_{ij} \qquad (8)$$

$$\sigma = \sqrt{\frac{1}{MN}\sum_{i=1}^{M}\sum_{j=1}^{N}(P_{ij}-\mu)^2} \qquad (9)$$

$$\theta = \frac{\sum_{i=1}^{M}\sum_{j=1}^{N}(P_{ij}-\mu)^3}{MN\sigma^3} \qquad (10)$$

$$\gamma = \frac{\sum_{i=1}^{M}\sum_{j=1}^{N}(P_{ij}-\mu)^4}{MN\sigma^4} \qquad (11)$$

M and N are the dimension of image. $P_{ij}$ is values of color on column $i_{th}$ and row $j_{th}$. For example, the two leaves in Fig. 4 have mean, standard deviation, skewness, and kurtosis as shown on Table 2.

TABLE 2
FEATURES OF LEAVES

| Features | Leaf 1 | Leaf 2 |
|---|---|---|
| R | $\mu$ = 72<br>$\sigma$ = 22.4763<br>$\theta$ = 5.2656e-005<br>$\gamma$ = -2.9997 | $\mu$ = 72<br>$\sigma$ = 26.1993<br>$\theta$ = 1.8794e-004<br>$\gamma$ = -2.9993 |
| G | $\mu$ = 161<br>$\sigma$ = 15.1716<br>$\theta$ = 7.8869e-005<br>$\gamma$ = -2.9997 | $\mu$ = 161<br>$\sigma$ = 20.1768<br>$\theta$ = 1.4060e-004<br>$\gamma$ = -2.9995 |
| B | $\mu$ = 31<br>$\sigma$ = 21.8473<br>$\theta$ = 4.2772e-005<br>$\gamma$ = -2.9998 | $\mu$ =<br>$\sigma$ = 24.1847<br>$\theta$ = 2.2727e-004<br>$\gamma$ = -2.9990 |

### C. Vein Features

Vein features can be extracted by using morphological opening [2]. That operation is performed on the gray scale image with flat, disk-shaped structuring element of radius--for example--1, 2, 3 and subtracted remained image by the margin. Based on that vein, 3 features are calculated as follow:

$$V_1 = \frac{A_1}{A}, V_2 = \frac{A2}{A}, V_3 = \frac{A_3}{A} \qquad (12)$$

In this case, $V_1$, $V_2$, and $V_3$ represent features of the vein, $A_1$, $A_2$, and $A_3$ represent total pixels of the vein, and A denotes total pixels on the part of the leaf.

### D. Texture Features

According to Petrou & Sevilla [17], The fractal dimension is not a good texture descriptor. Images are not really fractals, i.e. they do not exhibit the same structure at all scales. However, there is a fractal measure called lacunarity, which may help distinguish between two fractals with the same fractal dimension. Definitions lacunarity are shown as follows.

$$L_s = \frac{\frac{1}{MN}\sum_{m=1}^{M}\sum_{n=1}^{N}P_{mn}{}^2}{\left(\frac{1}{MN}\sum_{k=1}^{M}\sum_{l=1}^{N}P_{kl}\right)^2} - 1 \qquad (13)$$

$$L_a = \frac{1}{MN}\sum_{m=1}^{M}\sum_{n=1}^{N}\left|\frac{P_{mn}}{\frac{1}{MN}\sum_{k=1}^{M}\sum_{l=1}^{N}P_{kl}} - 1\right| \qquad (14)$$

$$L_p = \left(\frac{1}{MN}\sum_{m=1}^{M}\sum_{n=1}^{N}\left(\frac{P_{mn}}{\frac{1}{MN}\sum_{k=1}^{M}\sum_{l=1}^{N}P_{kl}} - 1\right)^p\right)^{1/p} \qquad (15)$$

Originally, those formulas are applied to gray scale images, where $P_{mn}$ is a grey value at coordinate (m,n) [18]. However, in our implementation, only Eq. 15 used as features with p equals 2, 4, and 6 and $P_{mn}$ represents value of color R, G, B of RGB image and intensity in grey scale image Therefore, there are 12 features to represent texture features.

*E. Feature Normalization*

Data normalization is a useful step often adopted, prior to designing a classifier, as a precaution when the feature values vary in different dynamic ranges [7]. In the absence of normalization, features with large values have a stronger influence on the cost function in designing the classifier. By normalizing data, value of all features will be in predetermined ranges. Normalization can be done by using formula as follow.

$$\hat{x}_i = \frac{x_i - x_{\min}}{x_{\max} - x_{\min}} \qquad (16)$$

In this case, $\hat{x}_i$ represents new value of the feature, $x_i$ represents original value of the feature, $x_{min}$ is the smallest value of original feature, and $x_{max}$ is the smallest value of original feature.

### III. PROPOSED SYSTEM

Proposed system for leaf classification is shown in Fig. 5. First, an image of the leaf is inputted into the system for classification. The features contained in the leaf are extracted by Feature Extractor. Then, the features are processed by a PNN. The result is an index that represents a plant. Then, Plant Information Getter translates the index into the name of the plant. Of course, before classification is done, the PNN has been trained once.

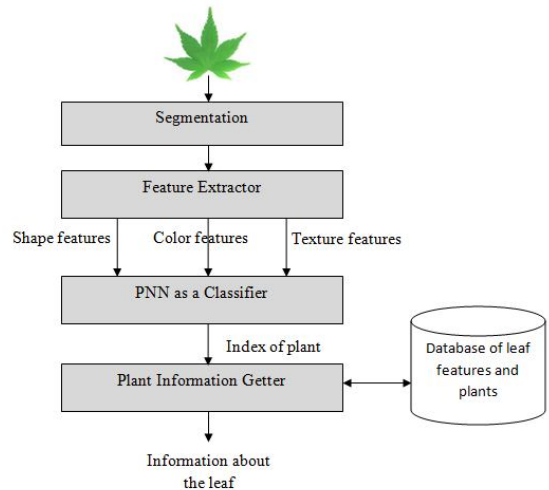

Fig. 5 Schema of the identification system

Segmentation is used to separate leaf from its background. It can be accomplished by using adaptive threshold as was used by Pahalawatta [27]. Firstly, an intensity histogram of image is built with 20 bins. Secondly, two major peaks in the histogram that represent the leaf and its background respectively are obtained. Third, find a bin with the smallest value that lies between the two major peaks. Then, the median of the bin is used as a threshold to separate leaf and its background.

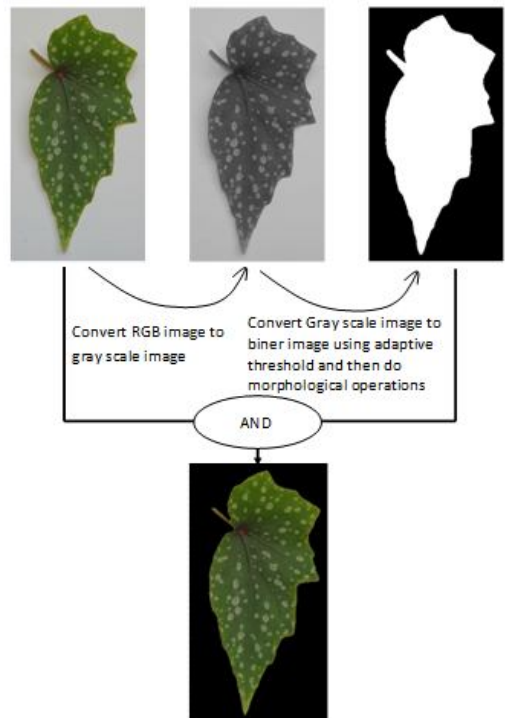

Fig. 6 Process to get area of the leaf

Fig, 6 illustrates process to get area of the leaf. Firstly, the image of the leaf is converted to gray level. Then, conversion from gray scale to binary form is done by using adaptive threshold. After that, several morphological operations are performed to remove holes in the leaf caused by previous thresholding. The leaf is obtained by using operation AND between RGB image and binary image.

The other important part of the identification system is PNN as a classifier. PNN is actually a kind of Radial Basis Function (RBF) which was proposed by Specht in 1989 [19]. Basically, PNN classifier adopts Bayes Classification rule and density estimation based on Gaussian function, that was proposed by Parzen. Fig. 10 shows its architecture.

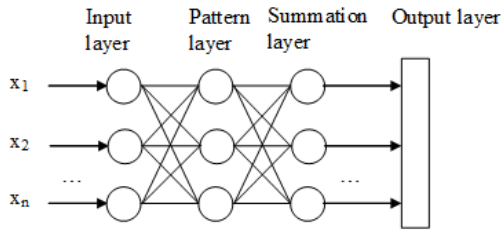

Fig. 7 Architecture of PNN

The input layer accepts an input vector. The pattern layer processed the input vector by using weight vector came from training dataset. This layer compute the distances from the input layer to the training input. As a result, a vector that indicates how close the input is to a training input. Then, in the Summation layer, a vector contains probabilities is found by summing up the contributions for each class. This vector of probabilities is sent to the output layer. The last layer in PNN structure produces a classification decision, in which a class with maximum probabilities will be assigned by 1 and other classes will be assigned by 0.

Mathematically, the probability is calculated by using Parzen method as follow.

$$p(x \mid w_j) = \frac{1}{(2\pi)^{d/2} \sigma^d n_j} \sum_{k=1}^{n_j} \exp\left(-\frac{(x - X_k)^2}{2\sigma^2}\right) \quad (17)$$

where $p(x \mid w_j)$ represents the conditional probability x to class $w_j$, $x$ is input vector, $X_k$ is training dataset, $d$ is number of the input vector, $n_j$ is number of samples for class j, $\sigma$ is smoothing factor that its value is inputted heuristically [28].

Based on fact that $x$ has class $j$ if

$$p(x \mid w_j) > p(x \mid w_j), \text{ for } i \neq j \quad (18)$$

then $p(x \mid w_j)$ can be calculated as follows [29]:

$$p(x \mid w_j) = \frac{1}{n_j} \sum_{k=1}^{n_j} \exp\left(-\frac{(x - X_k)^2}{2\sigma^2}\right) \quad (19)$$

## IV. EXPERIMENTAL RESULTS

To test the proposed method, a dataset called Flavia, that can be downloaded from http://flavia.sourceforge.net/, has been used. This dataset contain 32 kinds of plants.

Based on the dataset, 40 plants per species were used to train the network, and 10 plants per species were used to test performance of the system. In this case, PNN classifier is adjusted by using smoothing factor as equal 0.05.

In order to obtain performance of the system, the following formula [6] was used:

$$Performance = \frac{n_r}{n_t} \quad (20)$$

where $n_r$ is relevant number of images and $n_t$ is the total number of query.

Table 3 shows the results. As shown in the table, combination of shape, color (without kurtosis), vein, and texture features gives the best result with accuracy of 93.75%. Based on the results, we can see that all kinds of features have important contributions, except the kurtosis.

TABLE 3
VARIOUS FEATURES AND ITS PERFORMANCES

| Features | Performance |
|---|---|
| PFT | 74.6875% |
| PFT + 3 geometric features | 77.5000% |
| PFT + 3 geometric features + mean of colors | 82.5000% |
| PFT + 3 geometric features + mean of colors + standard deviation of colors | 88.1250% |
| PFT + 3 geometric features + mean of colors + standard deviation of colors + skewness of colors | 88.7500% |
| PFT + 3 geometric features + mean of colors + standard deviation of colors + skewness of colors + kurtosis of colors | 87.8125% |
| PFT + 3 geometric features + mean of colors + standard deviation of colors + skewness of colors + kurtosis of colors + 12 texture features | 90.6250% |
| PFT + 3 geometric features + mean of colors + standard deviation of colors + skewness of colors + 12 texture features | 90.0000% |
| PFT + 3 geometric features + 12 texture features | 85.3125% |
| PFT + 3 geometric features + mean of colors + standard deviation of colors + skewness of colors + 12 texture features + 3 vein features | 93.7500% |
| PFT + 3 geometric features + mean of colors + standard deviation of colors + skewness of colors + kurtosis of colors + 12 texture features + 3 vein features | 93.4375% |

## V. CONCLUSIONS

A method for leaf classification has been developed. The method incorporates shape and vein, color, and texture features and uses PNN as a classifier. Fourier descriptors, slimness ratio, roundness ratio, and dispersion are used to represent shape features. Color moments that consist of mean, standard deviation, and skewness are used to represent color. Twelve textures features are extracted from lacunarity. The result gives 93.75% of accuracy, which is slightly better than the original work that gives 90,312% of accuracy.

Although performance of the system is good enough, we believe that the performance still can be improved. Hence, other features will be researched in the future.